\documentclass{article}
\usepackage{ijcai20}

\usepackage{times}
\usepackage{soul}
\usepackage{url}
\usepackage[hidelinks]{hyperref}
\usepackage[utf8]{inputenc}
\usepackage[small]{caption}
\usepackage{graphicx}
\usepackage{amsmath}
\usepackage{amsthm}
\usepackage{booktabs}
\usepackage{multirow}
\usepackage{algorithm}
\usepackage{algorithmic}
\usepackage{subfigure}
\usepackage{textcomp}
\usepackage{latexsym} 
\usepackage{amsfonts}

\title{Intelligent Credit Limit Management in Consumer Loans \\ Based on Causal Inference}

\author{
Hang Miao, Kui Zhao, Zhun Wang, Linbo Jiang, Quanhui Jia, Yanming Fang, Quan Yu
\affiliations
Ant Financial Services Group\\
\emails
\{miaohang.mh, zhaokui.zk, huaishi.wz, xiaobo.jlb, quanhui.jia, yanming.fym, jingmin.yq\}@antfin.com\\
}

\begin{document}

\maketitle
\newcommand{\bigCI}{\mathrel{\text{\scalebox{1.07}{$\perp\mkern-10mu\perp$}}}}

\begin{abstract}
Nowadays consumer loan plays an important role in promoting the economic growth, and credit cards are the most popular consumer loan. One of the most essential parts in credit cards is the credit limit management. Traditionally, credit limits are adjusted based on limited heuristic strategies, which are developed by experienced professionals. In this paper, we present a data-driven approach to manage the credit limit intelligently. Firstly, a conditional independence testing is conducted to acquire the data for building models. Based on these testing data, a response model is then built to measure the heterogeneous treatment effect of increasing credit limits (i.e. treatments) for different customers, who are depicted by several control variables (i.e. features). In order to incorporate the diminishing marginal effect, a carefully selected log transformation is introduced to the treatment variable. Moreover, the model's capability can be further enhanced by applying a non-linear transformation on features via GBDT encoding. Finally, a well-designed metric is proposed to properly measure the performances of compared methods. The experimental results demonstrate the effectiveness of the proposed approach. 

\end{abstract}

\section{Introduction}
\label{section:intro}
Consumer loan helps people to finance their consumption demand, 
and nowadays it plays an important role in promoting the economic growth \cite{rona2018consumer}. 
The most popular consumer loans are credit cards, which enable borrowers 
to make everyday purchases \cite{hodson2014credit}. The credit limit (or line) management 
is one of the most essential parts in the risk management of credit cards. 
In credit limit management, a maximum loan amount is set for each credit account, 
according to customer's credit risk, consumer demand and so on. 
The total amount of money that customers own to the lender is referred as credit balance. 
Most lenders are trying to maximize their credit balance, 
since the balance is the foundation of their revenues. 
Through better credit limit management, the whole credit limit can be allocated to right places 
and the total credit balance can be then increased efficiently. 
For instance, a lender may give the opportunity of increasing the credit limit to a customer whose utilization is already very high as long as the credit risk is low enough. 
What's more, improving the credit limit management can also lead to better customer relationships, 
since more customers' consumption demands can be satisfied. 

\begin{figure}[!t]
\centering
\centerline{\includegraphics[width=3.1in]{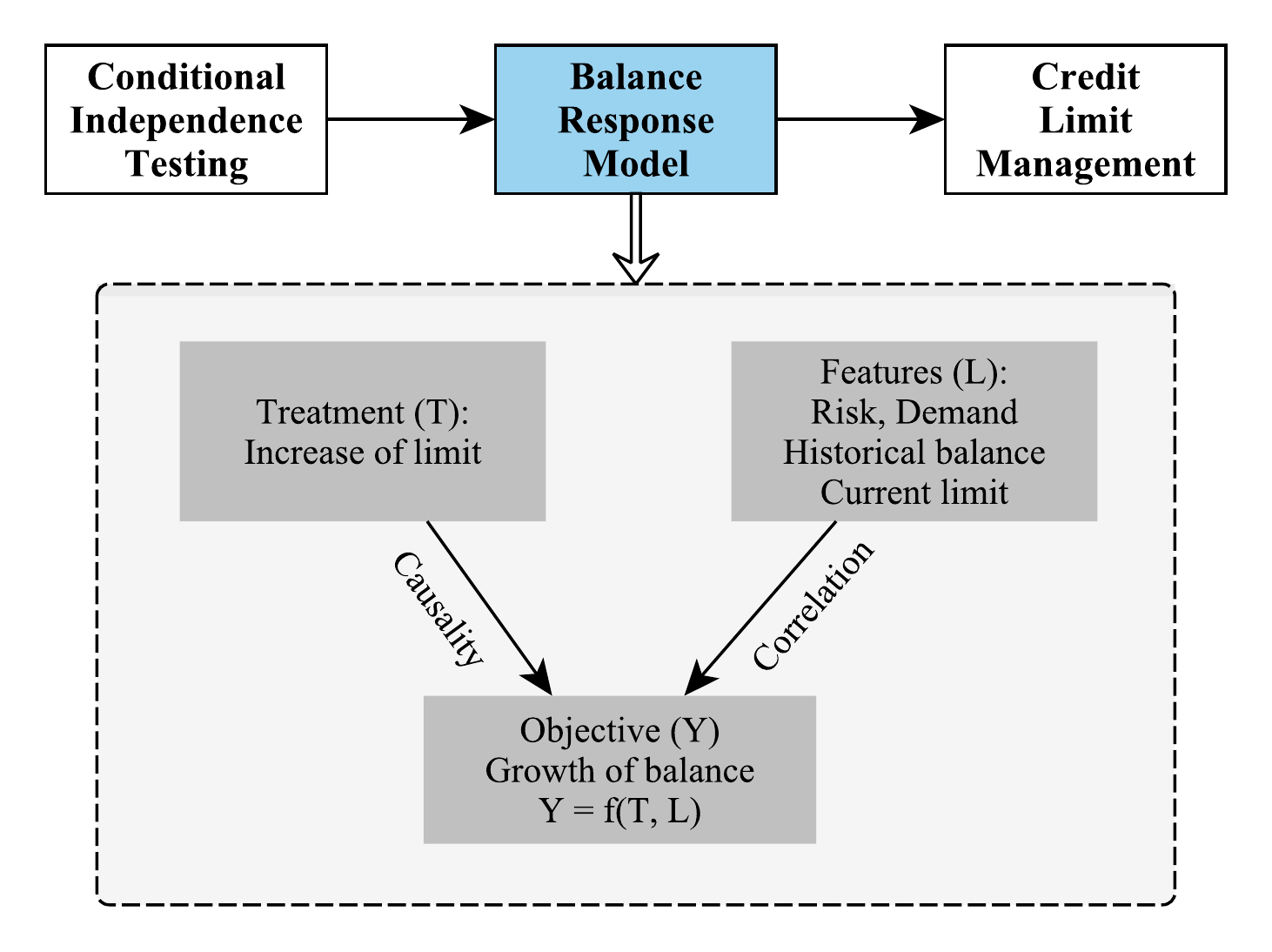}}
\caption{Our framework of intelligent credit limit management.}
\label{fig:framework}
\end{figure}

When managing the credit limit, there are several factors to take into account, 
including credit risk, consumer demand, historical balance and current limit etc. 
In traditional credit limit management, the limit is adjusted by experienced professionals 
in a heuristic and rule-based way. Although quantitative methods have been already utilized 
in credit limit management for a long time \cite{rosenberg1994quantitative} 
\cite{gross2000consumer} \cite{sohn2014optimization}, 
the whole process heavily relies on manual analyses and interventions, 
which greatly limits the sophistication of limit adjusting strategies. 
More specifically, the customers can only be segmented into a few heterogeneous 
groups, and the decision of limit adjustment is made for each 
group according to expert experiences. As the requirement of associated expert experiences 
is high, the traditional approach of managing credit limit is quite expensive. 
In addition, the development cycle of strategies is too long and 
the strategies is too simply for the dynamic and complex business environment 
in consumer loans nowadays. 

In recent decade, as many data about customers' behaviors have been collected, 
a lot of data-driven methods are leveraged to gain deeper insights into 
business environment, which helps us to make more informed decisions. 
In this paper, we present a data-driven approach to manage the credit limit 
in an intelligent way, and the whole framework is illustrated in Figure \ref{fig:framework}. 
To determine the increasing amount of credit limit for a particular customer, 
we need to predict the customer's response to the limit adjustment. 
In other words, we need to build a balance response model, 
which depicts the relationship between the increase of limit (i.e. treatment) 
and the growth of balance (i.e. objective). Different customers have different 
balance response curves, and the shape of each curve is determined by 
the specific factors (i.e. control variables or features) of corresponding customer. 
Most traditional machine learning methods are based on correlation study. 
They are incompetent to model the reliable relationship between the 
treatment and the objective, and thus cannot generalize to counterfactual 
predictions\footnote{The empirical evidence on real-world data will  
be later illustrated in the experimental section.}. 
Recently, causal inference attracts a lot of attentions from both research and industry 
communities. A more robust and stable decision-making system can be built 
if we model causality instead of correlation between the decision and the outcome 
\cite{hernan2010causal} \cite{imbens2015causal}. 
 
The majority of existing researches on causality focus on modeling with observed data, 
where treatments are not assigned randomly along control variables but 
intervened by some unknown confounders. The observational study relies on several assumptions, 
and it might be hard to validate them in practice, especially for the 
conditional independence assumption and the common support assumption \cite{peters2017elements}. 
On the other hand, although randomized testing is the {\it gold standard} 
for estimating the causal effect, it is costly and even infeasible in most scenarios. 
To overcome these challenges, we design and conduct a conditional independence testing, 
where the above two assumptions are perfectly satisfied, to acquire the data for building models.  
Based on the testing data, a structural outcome regression model is built to measure 
the heterogeneous treatment effect of increasing credit limits for different customers. 
The diminishing marginal effect is a common phenomenon in many scenarios and 
also exists in credit limit management without surprise. We introduce a carefully 
selected log transformation to the treatment variable to incorporate this prior knowledge. 
Furthermore, inspired by \cite{he2014practical}, the capability of our structural model 
can be enhanced by applying a non-linear transformation on features via GBDT encoding. 
Finally, we propose a proper evaluation metric to measure the performances of compared methods. 
The experimental results demonstrate the effectiveness of our approach. 

The rest of our paper is organized as follows. We briefly review related work in Section 2. 
We present the setup of testing in Section 3, and how to build and evaluate the balance response 
model in Section 4. The experimental results and further analyses are presented in Section 5. 
The conclusions and future plans are given in Section 6. 

\section{Related Work}
\label{section:rw}
The credit card is the most popular type of consumer loan \cite{hodson2014credit}, 
and quantitative methods have already been utilized in its risk management for a long 
time \cite{rosenberg1994quantitative} \cite{thomas2000survey}. 
One essential part in the risk management of credit cards is credit limit management. 
\cite{soman2002effect} argued that consumers use their credit limits as a signal of 
future potential earnings, and hence the credit limit would positively impact their spend. 
\cite{gross2000consumer} and \cite{song2011study} investigated consumers' 
responses to the change of credit limits, and the results show that increases 
in credit limits can effectively raise the credit balance. Inspired by these results, 
the credit limit can be adjusted by experienced professionals in a simple heuristic and rule-based way. 
To increase the sophistication of credit limit management, 
some more advanced approaches have been proposed. 
\cite{dey2010credit} discussed the possibility of using simulation along with action-effect models
to set the optimum credit limit for each account. 
\cite{paul2017consumer} tried to assign credit limit to new customers 
using Bayesian decision theory and Fuzzy logic. Unfortunately, these works focus on 
concepts or theoretical studies without empirical results, 
and thus it is hard to assess their effectiveness in real-world applications. 
\cite{sohn2014optimization} developed a strategy to maximize the total 
net profit by adjusting individual credit limits, 
and they demonstrated its effectiveness on real-world data from FICO. 
However, the customers can only be segmented into a few heterogeneous groups, 
and the whole process heavily relies on manual analyses and interventions. 

In recent decade, data-driven methods have been successfully applied to 
many business applications \cite{jordan2015machine}, 
such as credit risk models in consumer loans \cite{khandani2010consumer} \cite{addo2018credit}. 
In order to manage the credit limit in a more sophisticated and intelligent way, 
the response of each customer to the limit adjustment need to be predicted. 
Although big data and powerful algorithms can achieve better forecasting performance 
in many tasks, there are gaps between forecasting and decision making \cite{athey2017beyond}. 
Traditional machine or deep learning methods, which are based on correlation study, 
are incompetent to model the causality between the input factors and the target. 
Recently, causal inference has been widely studied and applied to many 
domains \cite{hernan2010causal} \cite{pearl2009causality} \cite{morgan2015counterfactuals} 
\cite{imbens2015causal}. One important topic in causal inference 
is to estimate the average treatment effect \cite{jorda2016time} \cite{machado2019instrumental}. 
The other equally important topic is to estimate the heterogeneous treatment effect, 
where \cite{athey2016recursive} and \cite{wager2018estimation} 
are two popular methods under the binary treatment. 
In this work, we are interested in estimating the heterogeneous treatment effect 
of increases in the credit limit, which are continuous treatments, for different customers. 
Moreover, although the common observational study has several drawbacks, 
they can be readily overcome in practice by conducting a conditional 
independence testing \cite{peters2017elements}. 
To our best knowledge, we are the first to utilize the causal inference 
to tackle the credit limit management in the real-world scenario.

\begin{table*}[t]
\centering
\caption{Testing setups for sub-prime customers with different credit ratings.}
\label{table:sub-prime}
\begin{tabular}{|c|c c c c c c c c |}
\hline
\multirow{2}{*}{Credit rating}       & \multicolumn{8}{ c |}{Treatment}                  \\ \cline{2-9} 
                       		& 0      &  5	& 10	    & 20		&  30	  	  & 60		& 100		& 150		\\ \hline
Very Good   		& 20$\%$           &  		          	&   		             &    		        	&  20$\%$          	  & 20$\%$           	& 20$\%$          	& 20$\%$ 			\\ 
Good   	& 20$\%$           &  			        	& 		             & 20$\%$           	&  20$\%$          	  & 20$\%$           	& 20$\%$         		& 				\\ 
Fair  		& 20$\%$           &  		         	& 20$\%$   	    & 20$\%$           	&  20$\%$          	  & 20$\%$           	&		        		&				\\ 
Poor   	& 20$\%$           &  20$\%$          	& 20$\%$              & 20$\%$           	&  20$\%$          	  & 		           	& 	         		& 				\\ \hline 
\end{tabular}
\end{table*}

\section{Testing}
\label{sec:testing}
We are interested in predicting the potential outcome (i.e. growth of balance) 
under different treatments (i.e. increases of limit) when keeping everything else constant, 
and it is referred as counterfactual prediction. For example, we want to know 
the growth of balance for a customer if we increase her credit limit by 10000, 
which has not ever been observed in the real-world. Since the counterfactual prediction 
is beyond the ability of traditional machine learning methods, we turn to causal inference 
and build a balance response model to estimate the heterogenous treatment 
effect when increasing the credit limit for a customer. 
However, a fundamental problem in causal inference is that only one treatment could be conducted 
for an individual at the same time, which means the data for the counterfactual reasoning 
are missing \cite{hernan2010causal}. Based on the observed data, 
this problem can be solved in observational study 
if the bias caused by confounders can be effectively eliminated. 

\subsection{Observational study}

\begin{table}[t]
\centering
\caption{Testing setups for prime customers with high demand level.}
\label{table:prime_high}
\begin{tabular}{|c| c c c c|}
\hline
\multirow{2}{*}{Demand level}       & \multicolumn{4}{c|}{Treatment}                  \\ \cline{2-5} 
                       &  0           & 100    		& 200              	&  300          \\ \hline
High   & 25$\%$           	 	&  25$\%$          	& 25$\%$               & 25$\%$                \\ \hline
\end{tabular}
\end{table}

In the observed data, treatments were assigned through some unknown strategies. 
There may be some factors that affect both treatment assignment and outcome, 
and they are referred as confounders. Thus for observational study, a major challenge is 
how to eliminate the bias caused by confounders. Under the conditional independence 
assumption (or unconfoundedness), which assumes all confounders have been observed, 
several techniques can be exploited to overcome this challenge. 
A commonly used technique is based on the propensity score \cite{rosenbaum1983central}, 
which is the probability that a customer is given a specific treatment $t$:
\begin{equation}
  \text{score}(t, \mathbf{L}) := P(T = t | \mathbf{L}), 
\end{equation}
where $\mathbf{L}$ denotes the observed features, and $T$ denotes the treatment. 
As long as the propensity score is estimated, 
the Inverse Probability of Treatment Weighting (IPTW) 
can be leveraged to balance the distribution of treatments 
and remove the bias caused by confounders \cite{hirano2003efficient}:
\begin{equation}
\text{IPTW}(t, \mathbf{L}) := \frac{1}{\text{score}(t, \mathbf{L})}. 
\end{equation}
As we can see,  the IPTW method also implicitly relies on another assumption that 
there should exist overlaps among different treatments under the same observed features:
\begin{equation}
 0 < P(T| \mathbf{L}) < 1. 
\end{equation}
This assumption is also known as common support. 
Actually, for many other techniques in observational study, such as outcome regression, 
the above two assumptions are also required, though it might be hard to validate them in practice. 

\begin{table}[t]
\centering
 \caption{Testing setups for prime customers with low demand level.}
 \label{table:prime_low}
\begin{tabular}{|c|c c c c|}
\hline
\multirow{2}{*}{Demand level}       & \multicolumn{4}{c|}{Treatment}                  \\ \cline{2-5} 
                       &  0           &  50    		& 100           	& 150          \\ \hline
Low   & 25$\%$           	 	&  25$\%$          	& 25$\%$               & 25$\%$                \\ \hline

 \end{tabular}
\end{table}

\subsection{Conditional independence testing}
The randomized testing is the gold standard for causal inference, 
and the mentioned assumptions can be satisfied naturally \cite{imbens2015causal}. 
However, randomized testing is costly and even infeasible in most scenarios. 
For example, we cannot increase the credit limit too much 
for a customer who has a poor credit rating, because it is very likely to cause a loss. 
Fortunately, we can conduct a conditional independence testing that satisfies 
those assumptions as well \cite{peters2017elements}. In the conditional independence testing, 
treatments are randomly assigned conditional on specific features $\mathbf{Z} \subset \mathbf{L}$, 
and the dependence between features $\mathbf{L}$ and treatments $T$ is 
broken down under the given $\mathbf{Z}$: $T$ $\bigCI$ $\mathbf{L} | \mathbf{Z}$.
Since the confounders $\mathbf{Z}$ are introduced artificially, 
the conditional independence assumption is obviously satisfied. 

When designing the conditional independence testing, 
it is crucial to choose the proper artificial confounder. 
In this work, we construct $\mathbf{Z}$ based on the credit rating and the consumer demand. 
At first, we split the customers into two groups according to their credit rating: 
prime group and sub-prime group. For the sub-prime group, 
the customers are further segmented into four subgroups with different credit ratings: 
very good, good, fair, and poor (as illustrated in Table \ref{table:sub-prime}) \footnote{Without loss of generality, all the treatment values have been scaled by a constant to avoid divulging business secrets.}. 
For the prime group, the customers are further segmented into 
two subgroups with different demand levels:
high and low (as illustrated in Table \ref{table:prime_high} and Table \ref{table:prime_low}). 
The dummy coding of above six subgroups is set as $\mathbf{Z}$. What's more, 
to meet the common support assumption, we keep a proportion of customers' credit limits 
unchanged in each subgroup. As we can see, the increase of credit limit in each subgroup 
is bounded in a range, which is determined by business rules and will be maintained in productization. 
For instance, the increases of credit limits are relatively small for the customers with high credit risk, 
and greater increases of credit limits are assigned to the prime customers with higher demand level. 

\section{Balance Response Model}
\label{sec:response}
Based on the testing data, we build a balance response model to predict 
the growth of balance under increases in the credit limit for different customers. 
The model is built within the framework of potential outcome \cite{rubin2005causal} 
\cite{edin2018outcome}. 

\subsection{Problem formulation}
Firstly, let us formally introduce several important conceptions used in our model:
\\
{\bfseries Treatment:} Treatment $T$ is defined as the increase of credit limit for a customer, 
and it is a continuous variable. 
\\
{\bfseries Outcome:} Outcome $Y$ is defined as the growth of 
credit balance over a period of time, 
$Y(T)$ denotes the potential outcome under the treatment $T$. 
For example, if $B_{1}$ (and $B_{2}$) denotes the monthly average balance 
before (and after) increasing the credit limit by $T$, $Y(T) = B_{2} - B_1$. 
\\
{\bfseries Feature:} There are plenty of features in real-world productization, 
and we select some representative features in this paper without loss of generality. 
\begin{itemize}
\item Credit risk: the probability of default, the credit rating. 
\item Consumption: the monthly average spends over the last 3 months 
and 6 months respectively, the monthly maximum spend over the last 12 months. 
\item Historical balance: the monthly average balances over the last 3 months 
and 6 months respectively. 
\item Current limit: the credit limit before adjustment.
\end{itemize}
Besides these features, the artificial confounder $\mathbf{Z}$ should also 
be included in features $\mathbf{L}$. Then the outcome regression model 
is correctly specified, and no more adjustment is required \cite{hernan2010causal}. 

We aim to build a model $f(\cdot)$ to conduct the counterfactual 
prediction $Y(T)$ given the features $\mathbf{L}$ and treatment $T$:
\begin{equation}
Y(T) = f(T, \mathbf{L}) = \mathbb{E}[Y | \mathbf{L}, T].
\end{equation}
As the treatment $T$ is a continuous variable, the local effect around 
a treatment point can be calculated, and it is referred as heterogeneous marginal effect:
\begin{equation}
  \partial Y(T) /  \partial T = \mathbb{E}[\partial Y / \partial T | \mathbf{L}].
\end{equation}
The straightforward linear regression is the most simple choice to model 
the relationship between $\mathbf{L}$, $T$ and $Y$: 
\begin{equation}
\label{eq:lr}
 Y(T) =  \mathbf{L} \cdot \mathbf{w}_0 + T \times w_1. 
\end{equation}
The marginal effect can be directly interpreted by the coefficient $w_1$:
\begin{equation}
  \partial Y(T) /  \partial T =  w_{1}.
\end{equation}
As we can see, the marginal effect is independence of $\mathbf{L}$, 
which means that customers with different features would have the same marginal causal effect, 
and it is obviously inconsistent with the actual situation. 

\begin{figure}[!t]
\centerline{\includegraphics[width=3.2in]{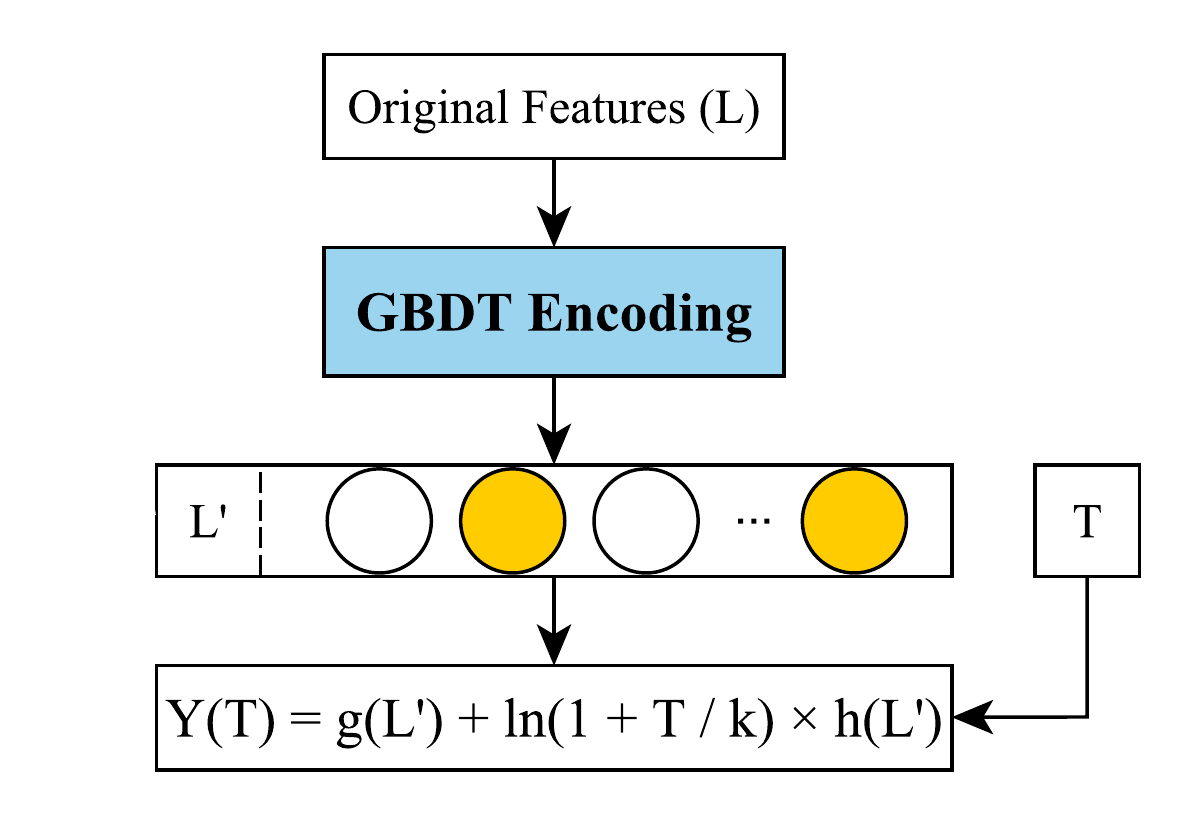}}
\caption{The structural outcome regression with both log transformation and GBRT encoding.}
\label{fig:model}
\end{figure}

\subsection{Structural outcome regression}
In fact, customers with different features should have different marginal causal effects. 
If we assume that the outcome $Y(T)$ is a linear function of the treatment $T$, 
the relationship between $\mathbf{L}$, $T$ and $Y$ can be depicted by 
a  structural outcome regression model: 
\begin{equation}
\label{eq:or}
Y(T) = g(\mathbf{L}) + T \times h(\mathbf{L}).
\end{equation}
Note that the simple linear regression is a special case, 
where $g(\mathbf{L}) = \mathbf{L} \cdot \mathbf{w}_0$ and $h(\mathbf{L}) = w_1$. 
The heterogeneous marginal effect is computed as follow:
\begin{equation}
  \partial Y(T) /  \partial T =  h(\mathbf{L}),
\end{equation}
which is determined only by features $\mathbf{L}$ and independent of the treatment $T$.

Actually, the diminishing marginal effect exists in the credit limit management, 
which means that the marginal causal effect 
should decrease along with the increase of treatment $T$. 
In order to incorporate this prior knowledge, 
we introduce a log transformation to the treatment as follow: 
\begin{equation}
\label{eq:orl}
Y(T) = g(\mathbf{L}) + \ln(1 + T / k) \times h(\mathbf{L}),
\end{equation}
where $k$ is a hyper-parameter. 
Then the marginal effect is computed as follow:
\begin{equation}
\partial Y(T) /  \partial T =  h(\mathbf{L}) / (T + k), 
\end{equation}
which indicates that the outcome $Y$ would increase slower as the treatment $T$ increases. 

In the traditional outcome regression, $g(\cdot)$ and $h(\cdot)$ are linear functions of $\mathbf{L}$, 
and their capabilities are limited. Inspired by \cite{he2014practical}, 
we apply a non-linear transformation to features $\mathbf{L}$ through a GBDT encoding. 
Firstly, a GBDT model is built to predict the outcome $Y$ based on features $\mathbf{L}$ 
without the treatment $T$. Then, the original features $\mathbf{L}$ 
is transformed into new features $\mathbf{L}'$ via the acquired GBDT model. 
Finally,  $\mathbf{L}'$ is used instead of $\mathbf{L}$ to build the outcome 
regression model as follow  (also illustrated in Figure \ref{fig:model}) :
\begin{equation}
\label{eq:gbdt_e_or_l}
Y(T) = g(\mathbf{L}') + \ln(1 + T / k) \times h(\mathbf{L}'). 
\end{equation}
When training the outcome regression model, 
the objective is to minimize the Mean Square Error (MSE), 
and the L1 or L2 regularization can be further utilized to enhance the model's generalization.  

\begin{figure*}[tbp!]
\centering
\subfigure[Ground truth]
{\includegraphics[width=1.90in, angle=0]{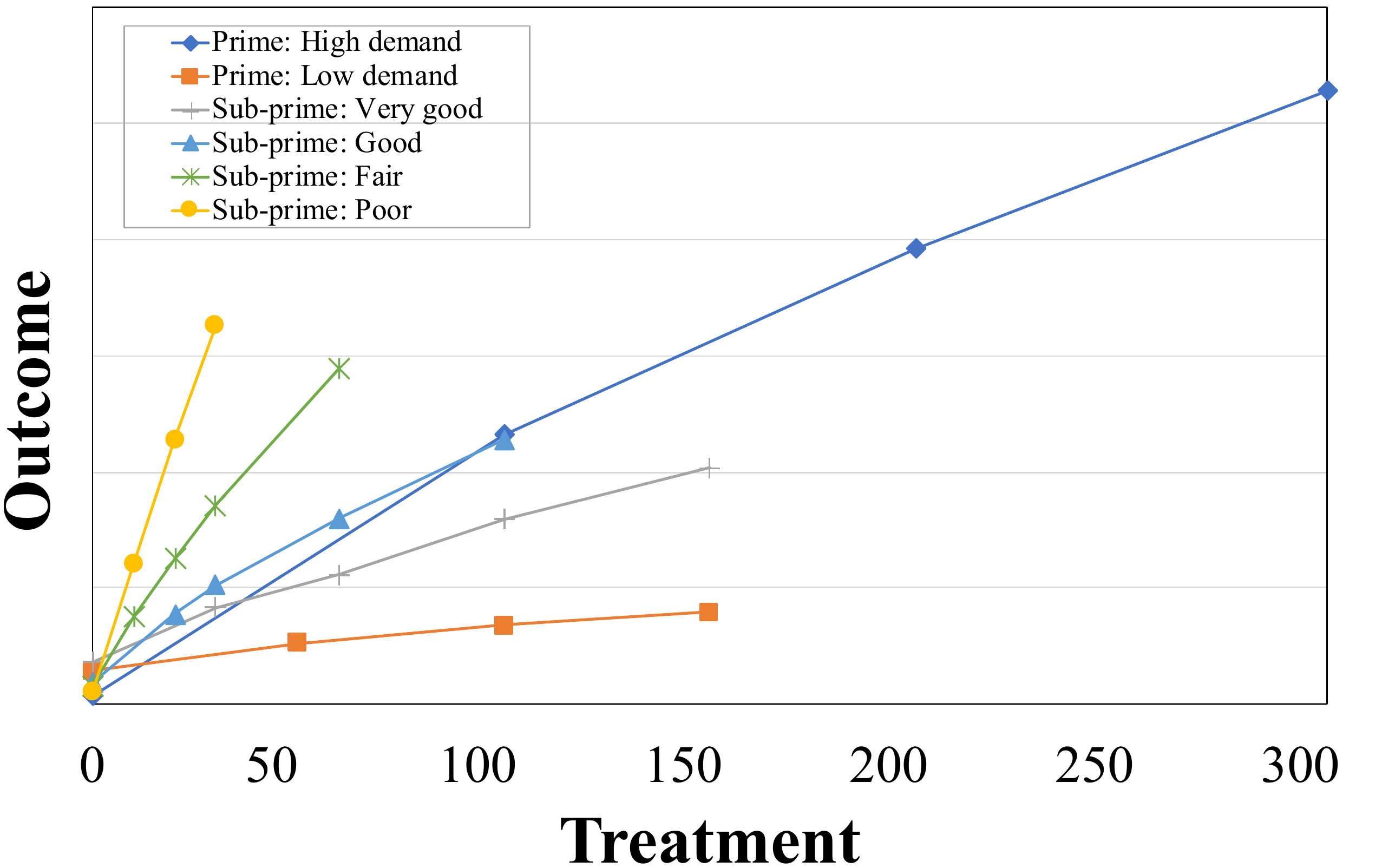}}
\subfigure[Single GBDT]
{\includegraphics[width=1.90in, angle=0]{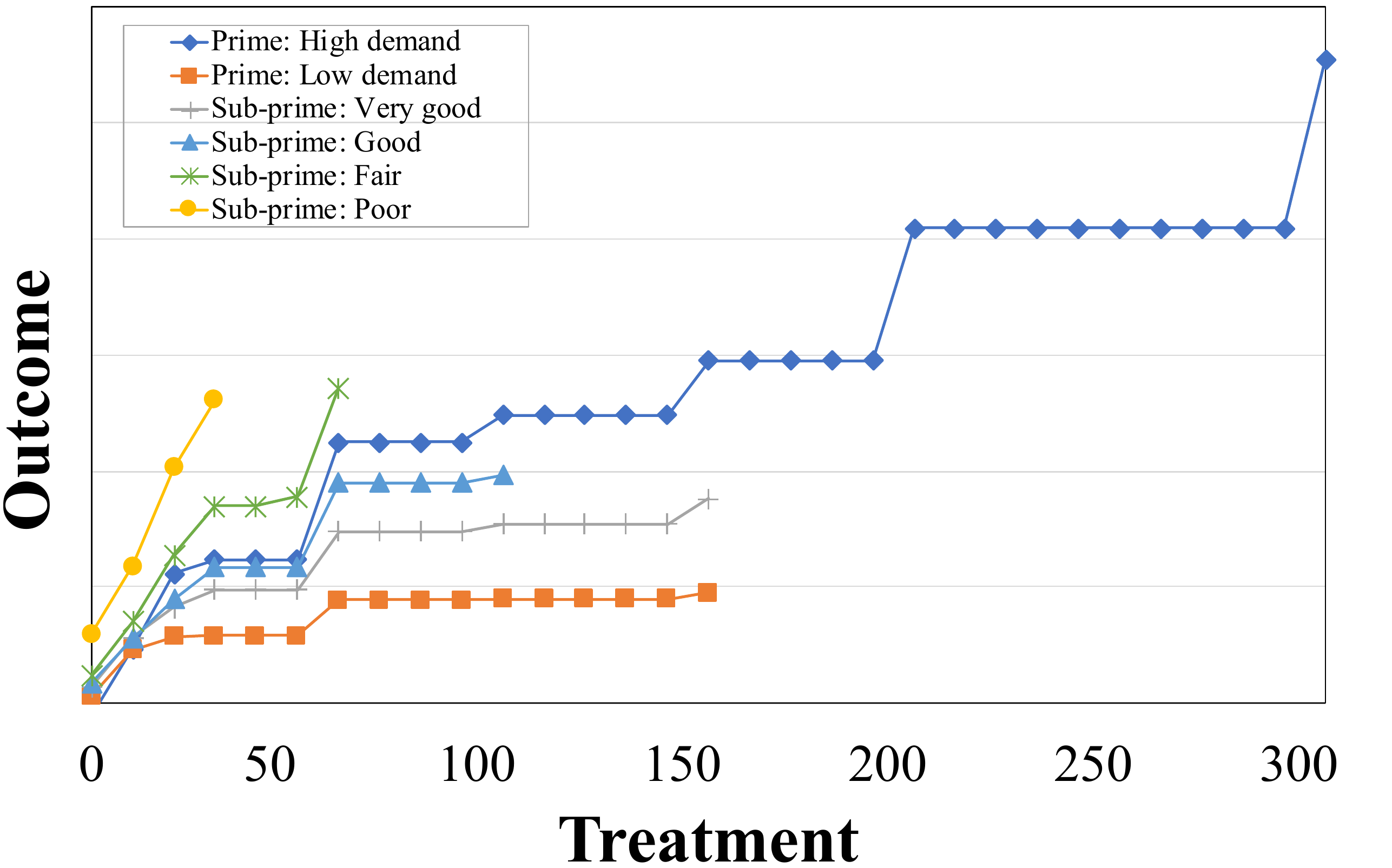}}
\subfigure[Linear regression]
{\includegraphics[width=1.90in, angle=0]{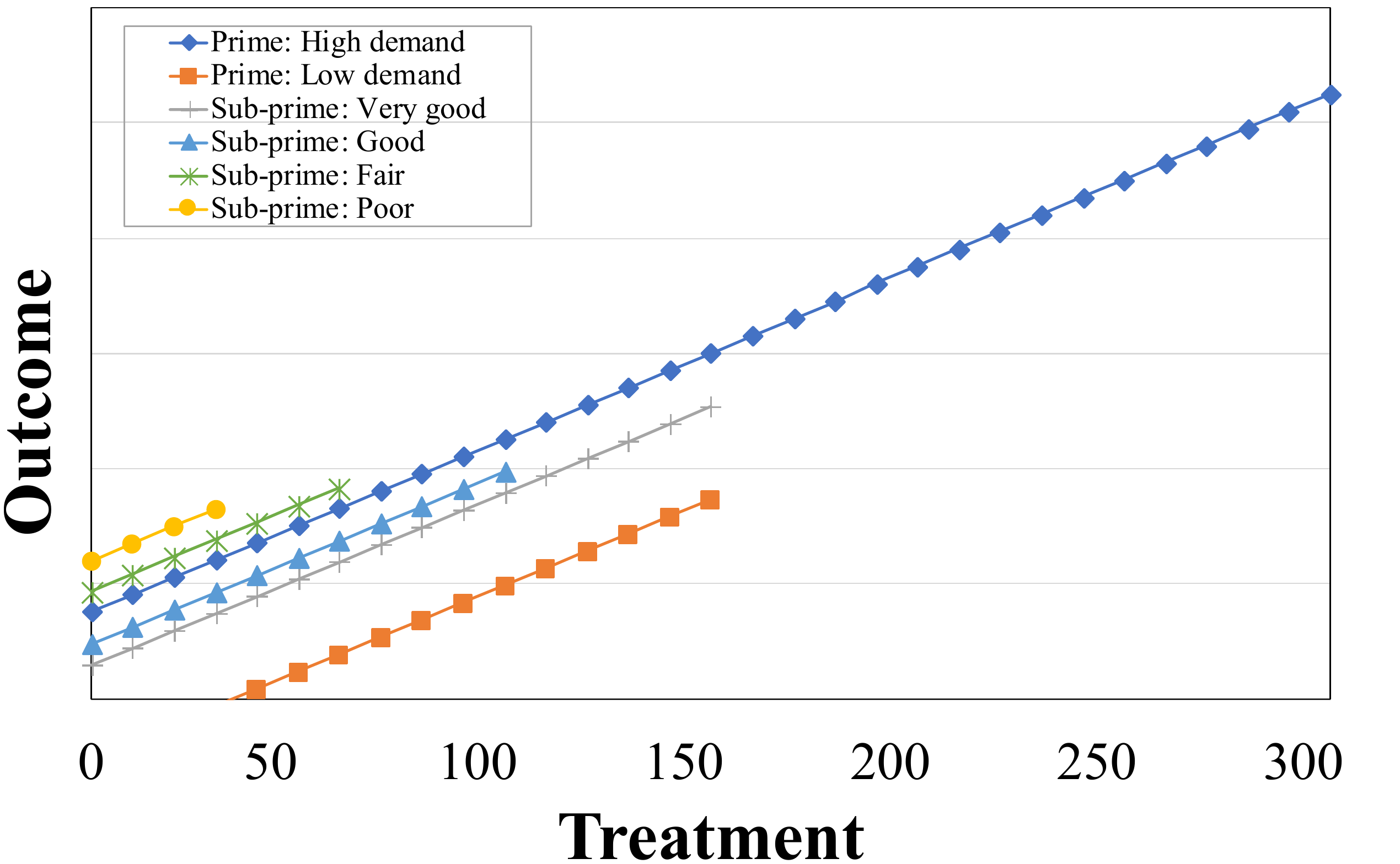}}
\subfigure[Outcome regression with $T$]
{\includegraphics[width=1.90in, angle=0]{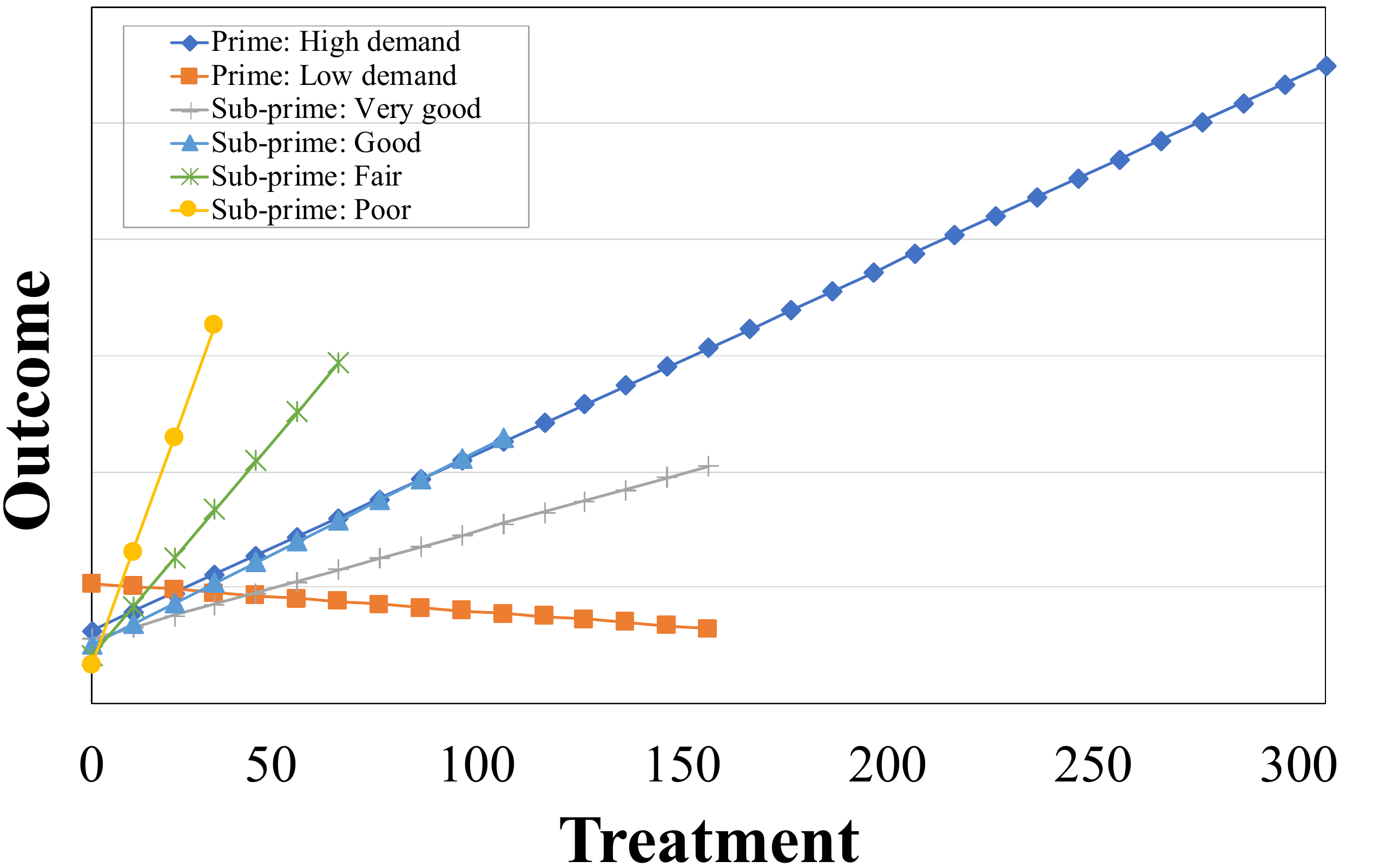}}
\subfigure[Outcome regression with $\ln(1 + T)$]
{\includegraphics[width=1.90in, angle=0]{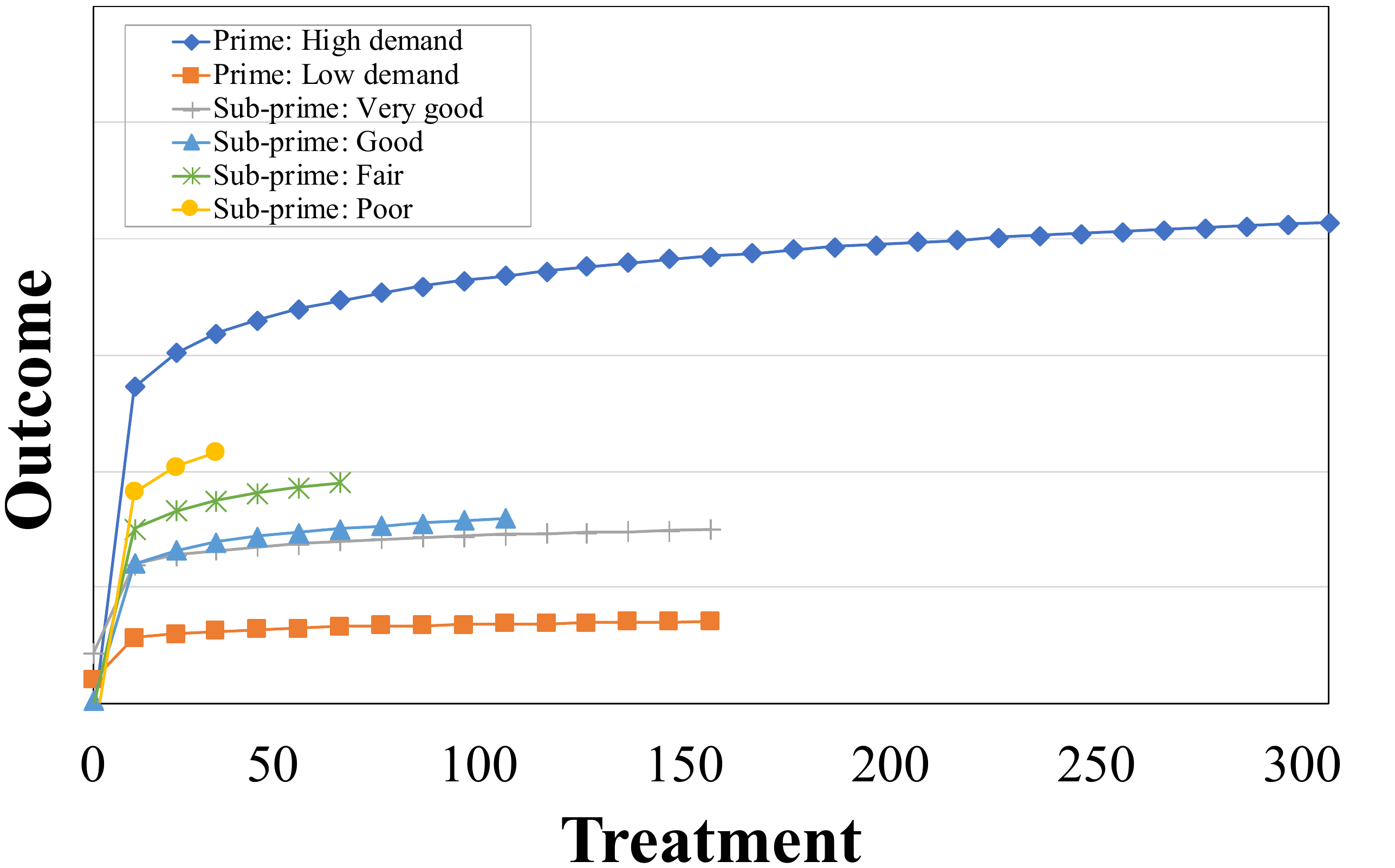}}
\subfigure[Outcome regression with $\ln(1 + \frac{T}{20000}$)]
{\includegraphics[width=1.90in, angle=0]{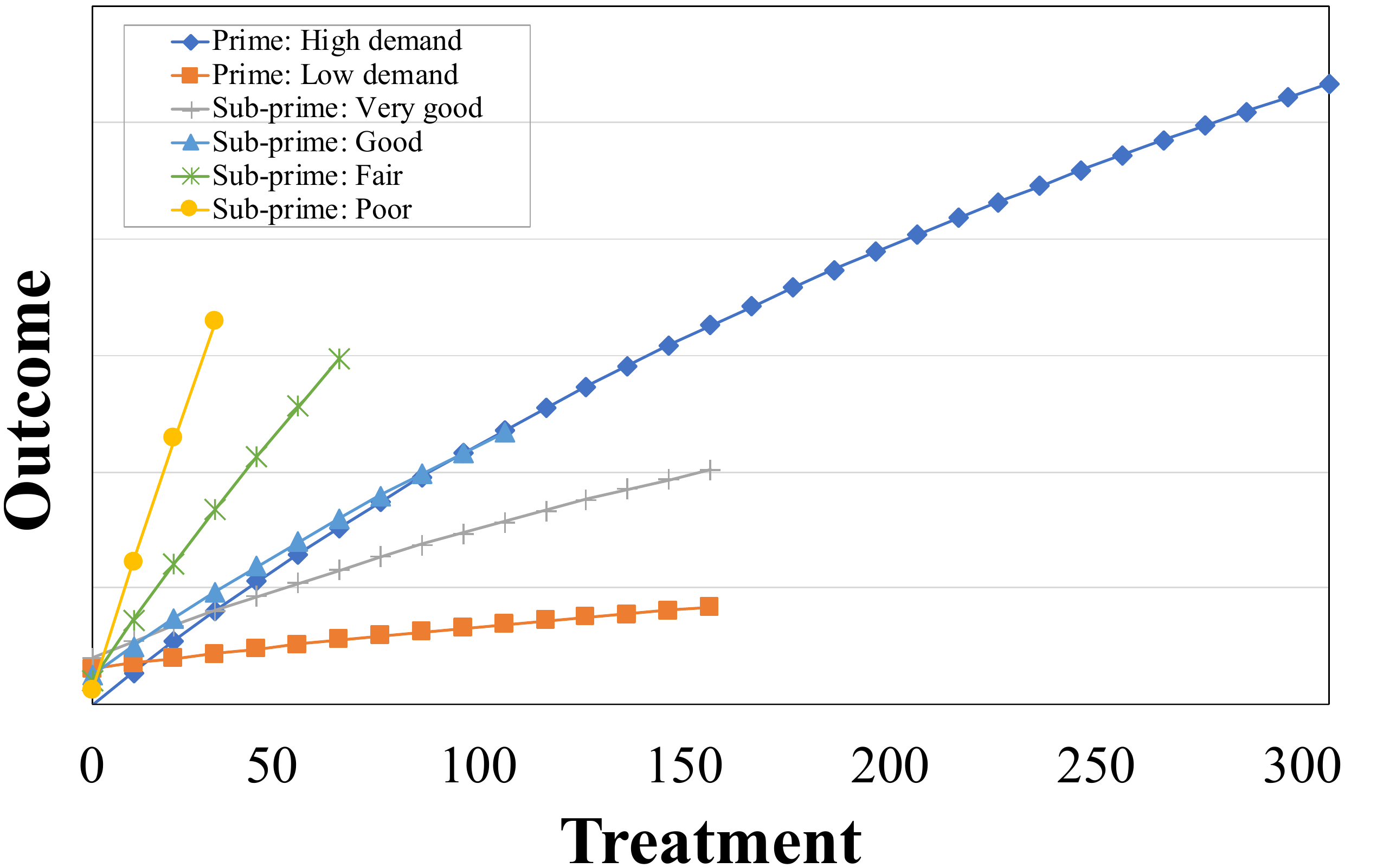}}
\caption{The average response curves of six subgroups 
in testing under different methods.}
\label{fig:vis}                
\end{figure*}

\subsection{Measurement of performance}
\label{sec:metric}
In practice, there exists great randomness in a customer's consumption, 
which brings strong uncertainty to the observed growth of balance for a specific customer. 
Therefore, the divergence between observed outcome and prediction is 
dominated by noises if we calculate the prediction error at the individual level. 
So a proper measurement is required to evaluate compared methods. 
If we assume that an observed outcome is the sum of the expected treatment effect 
and a gaussian noise centered at zero, it is possible to cancel out noises 
and obtain the expected treatment effect 
by averaging the observed outcomes of customers with the similar features. 
Based on that, we propose a novel method to measure the performance of different methods. 
Firstly, the customers are segmented into many heterogeneous groups according to their features, 
and the customers is homogeneous within each group. When computing the prediction error, 
the weight of each group is set as the number of customers in it. 
Then the Relative Mean Absolute Error (RMAE) is calculated as follow 
to measure the performance of compared methods:
\begin{equation}
\text{RMAE} = \frac{\sum_{i=1}^N {w_i \times |\hat{y}_{i} - y_{i}|}}{ \sum_{i=1}^N {w_i \times y_{i}}}, 
\end{equation}
where $N$ denotes the number of groups. For $i$-th ($i=1,\dots, N$) group, 
$y_i$ is its expected growth of balance, $\hat{y}_i$ is the average prediction result, 
$w_i$ is the corresponding weight. 

In this paper, the customers are segmented into more than 6000 groups 
according to several crucial features, including the probability of default, 
the current credit limit, the monthly average spend 
and balance over the last 6 months respectively. 
Note that the number of groups should be adjusted accordingly. 
If there are too few groups, the performance of different methods will be similar. 
In contrast, the noises cannot be effectively cancelled out if there are too many groups. 

\section{Experiment}
\label{section5}
The testing was conducted in one virtual credit card scenario, 
which is a service provided by one of the biggest FinTech companies in the world. 
There are several millions of samples collected from the testing, 
and we randomly choose 50\% samples as the training set and another 50\% samples for test\footnote{The dataset does not contain any Personal Identifiable Information and it is desensitized and encrypted. Adequate data protection was carried out during the experiment to prevent the risk of data copy leakage, and the dataset was destroyed after the experiment. The dataset is only used for academic research and does not represent any real business situation.}. 

\subsection{Setups}
All continuous features except the probability of default are scaled by standardization.  
The credit ratings are represented by the dummy coding. 
The compared methods are as follow: 
\begin{itemize}
\item {\bf Linear Regression}: The linear regression in Eq. (\ref{eq:lr}).
\item {\bf Single GBDT}: A GBDT model is built to predict the outcome Y by taking both $\mathbf{L}$ 
and $T$ as the input. 
\item {\bf Outcome Regression (+LOG) }: The outcome regression shown in Eq. (\ref{eq:or}). 
After introducing log transformation to the treatment $T$, we can get Eq. (\ref{eq:orl}). 
\item {\bf GBDT Encoding + OR (+LOG) (+L1 or L2)}: A GBDT encoding can be leveraged to enhance 
the model's capability, i.e. Eq. (\ref{eq:gbdt_e_or_l}). Moreover, L1 or L2 regularization 
can enhance the model’s generalization. 

\end{itemize}

All hyper-parameters are turned by cross validation on the training set. 
For both single GBDT and GBDT encoding, 
the number of trees is set to $50$ and the max depth is set to $3$. 
The $k$ in log transformation is set to $20000$, 
and the regularization coefficient is set to 100 for both L1 and L2. 

 \begin{figure*}[tbp]
\centering
\subfigure[Partition by probability of default]
{\includegraphics[width=1.94in, angle=0]{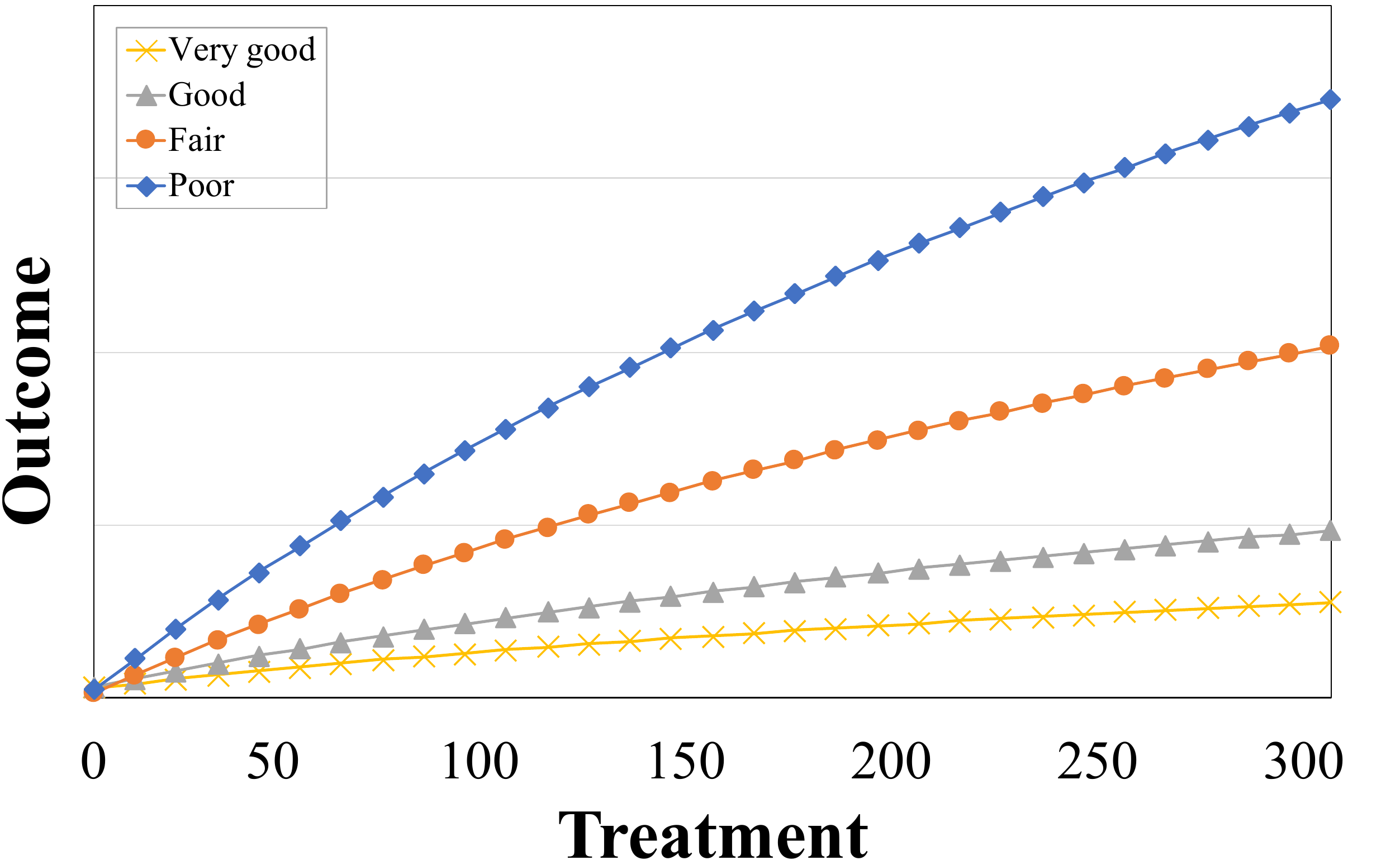}}
\subfigure[Partition by balance-to-limit ratio]
{\includegraphics[width=1.94in, angle=0]{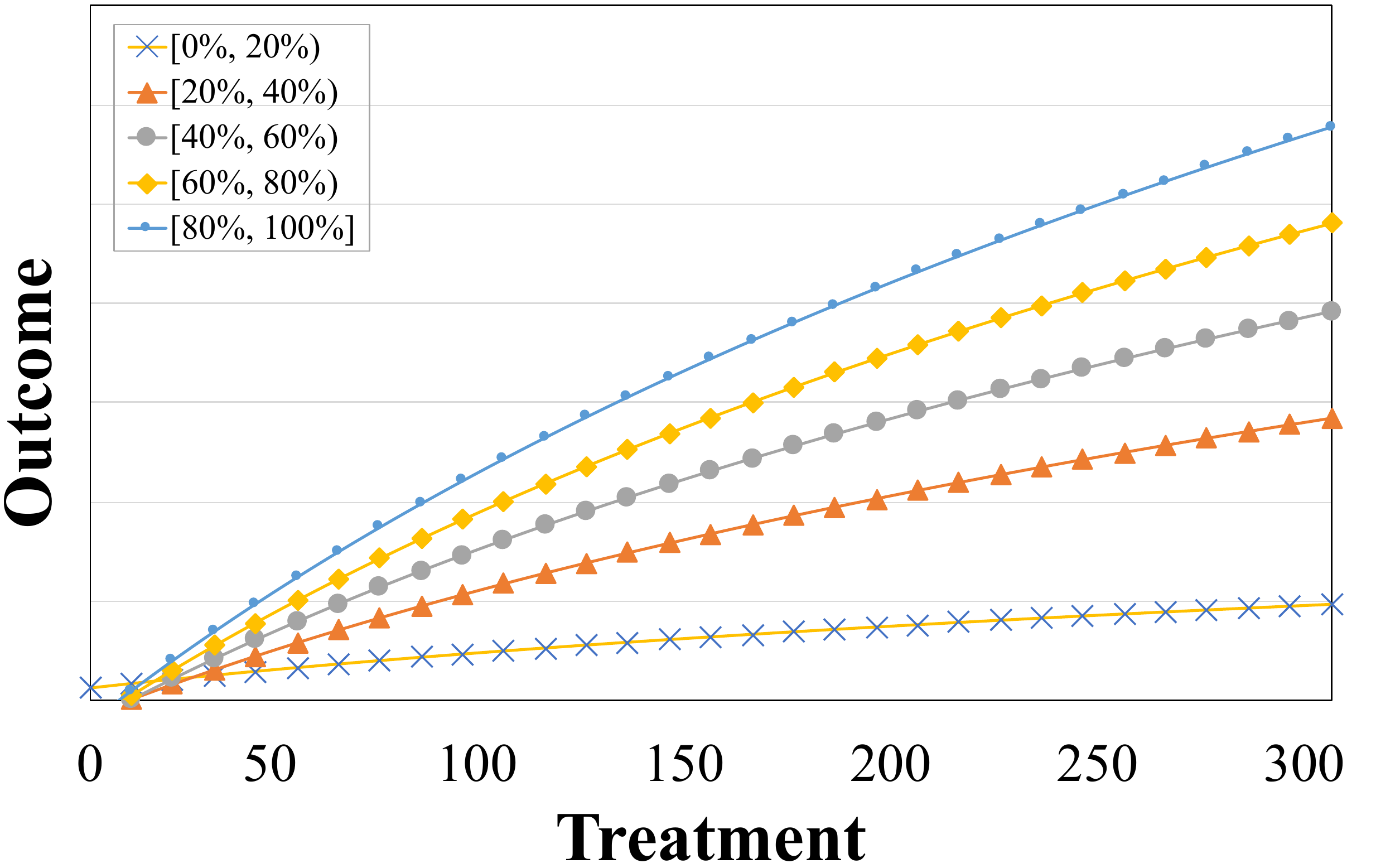}}
\subfigure[Partition by spend-to-limit ratio]
{\includegraphics[width=1.94in, angle=0]{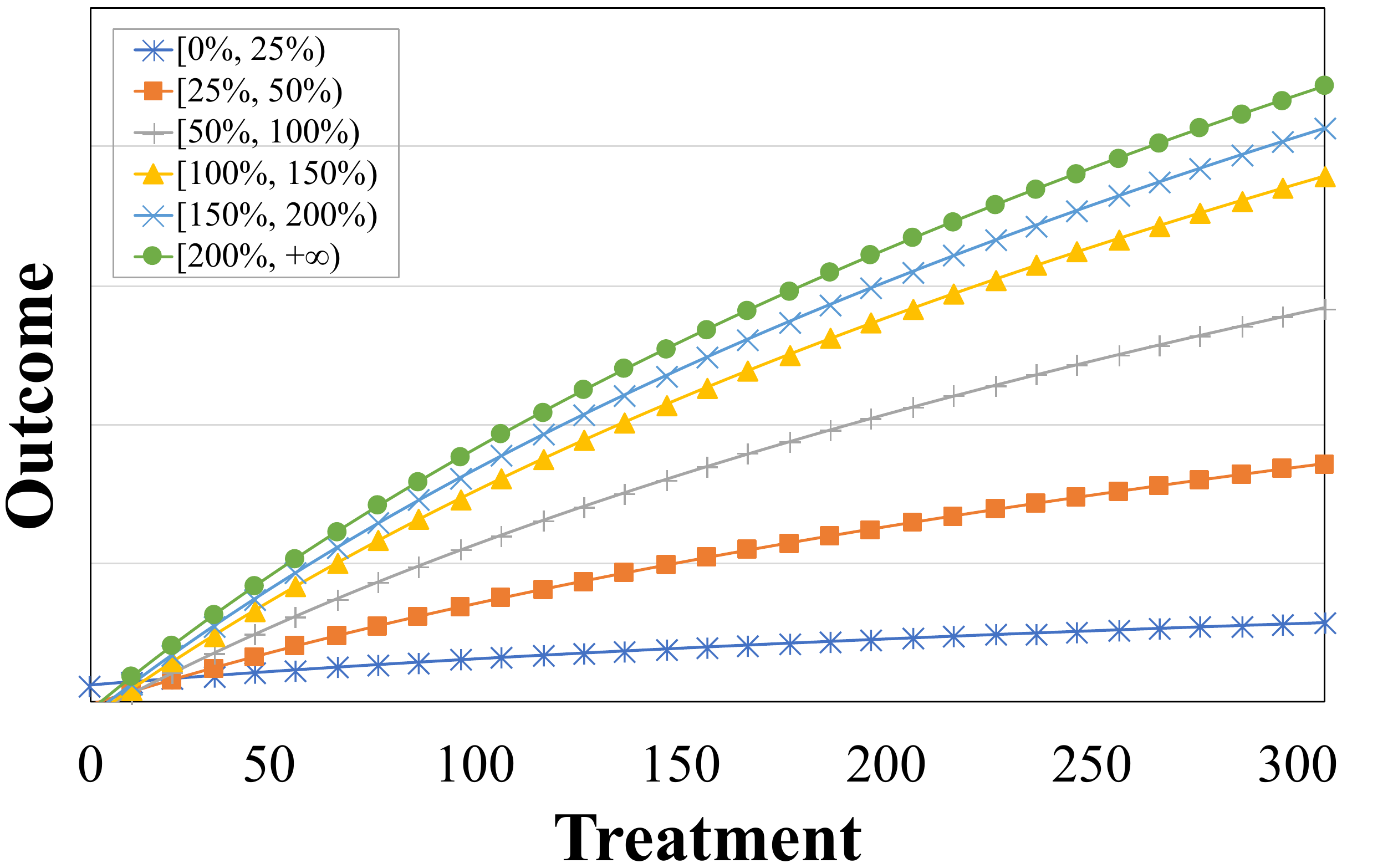}}
\caption{The partial dependence plots for different types of customers. }
\label{fig:group}                
\end{figure*}

\subsection{Results}
\begin{table}[h] 
\caption{\ The RMAE of different methods (lower is better). }
\label{table:result}
\centering
\begin{tabular}{l c c c}
\hline
\bfseries{Models} 									& \bfseries{Training} 	& \bfseries{Test}	\\
\hline
Linear Regression 									& 94.05\% 		& 94.10\%			\\
Single GBDT											& 47.44\% 		& 47.58\% 		\\
\hline						
Outcome Regression (OR)											& 87.05\% 		& 86.73\% 		\\
OR + LOG											& 85.28\% 		& 85.03\% 		\\
\hline
GBDT Encoding   + OR										& 49.31\% 		& 49.84\%			\\
GBDT Encoding   + OR + LOG								& 38.95\% 		& 39.35\% 		\\
GBDT Encoding   + OR  + LOG + L2 							& 38.88\%			& 39.26\% 		 \\
GBDT Encoding   + OR + LOG + L1							& {\bf 37.94}\%			&  {\bf 38.30}\%		\\
\hline
\end{tabular}
\end{table}

The results of all compared methods are reported in Table \ref{table:result}.
As we can see, the performance of simple linear regression is very poor, 
since customers with different features should have different marginal effects. 
Therefore, the outcome regression can effectively decrease the prediction error. 
After incorporating the diminishing marginal effect, the prediction error can be further decreased. 
With the help of GBDT encoding, 
the capability of model is significantly enhanced and the performance is improved dramatically. 
While both L1 and L2 regularizations can enhance the model’s generalization, 
the best result is achieved under the L1 regularization. 
It is probably because the new features from GBDT encoding are sparse, 
and L1 regularization is more suitable for that. 
Lastly, although the single GBDT can achieve competitive performance, 
it has nearly no ability of interpolation, 
which will be demonstrated with more details in the next subsection. 

\subsection{Analyses}
\subsubsection{Visualization of ablation.}
Now we visualize the prediction on six subgroups presented in Section \ref{sec:testing} to 
intuitively illustrate the motivations of outcome regression and log transformation. 
More specifically, for each customer in the test set, 
the outcomes are predicted by different methods under all treatments, 
ranging from 0 to the corresponding maximum value with a stride of 1000. 
After averaging the predicted results for each subgroup, 
the average response curves are shown in Figure \ref{fig:vis}. 
First of all, although GBDT is the most popular method in machine learning 
and its performance is outstanding in many tasks, 
it has nearly no ability of interpolation, and thus cannot generalize to counterfactual predictions. 
Different from the simple linear regression, the outcome regression allows 
customers with different features having different marginal effects. 
After introducing the log transformation to the treatment, 
the marginal effects are able to decrease along with the increase of treatments, 
and the intensity of diminishing marginal effect is controlled by the hyper-parameter $k$. 

\subsubsection{Partial dependence plots.}
As we have the response model for each customer, the average response curve for 
any specific type of customers can be easily obtained by averaging their treatment effects. 
By inspecting the Partial Dependence Plots (PDP) for different types of customers, 
we could gain deeper insights into the industry trends. 
Therefore, the response models can not only help to determine the adjustment of 
credit limit for a particular customer, but also help to make better macro-decisions. 
In this subsection, we partition the customers into several groups according to 
three most important factors, and the PDPs for different types of customers 
are presented in Figure \ref{fig:group}. 
As we can see, the marginal treatment effect is larger for customers with higher risk, 
probably because their current credit limits are quite low. 
For the customers with higher utilization or spends, it is likely that the current credit limits cannot 
cover their demands, and thus one unit increase of credit limit leads to higher growth of balance. 
All above observations are consistent with our expert experiences. 

\section{Conclusion}
\label{sec:conclustion}
In this paper, we propose a data-driven approach based on causal inference 
to manage the credit limit in intelligent ways. Firstly, we design and conduct 
a conditional independence testing. Based on the acquired data from testing, 
a structural outcome regression model is built to measure the heterogeneous 
treatment effect of increasing credit limits. 
Through a carefully selected log transformation, 
the diminishing marginal effect has been incorporated. The capability of 
our model is further enhanced by applying GBDT encoding to features. 
Finally, we propose a proper evaluation metric to measure 
the performances of compared methods. The experimental results confirm 
the effectiveness of our approach and the analyses are inspiring.

For future works, 
there is a strong motivation to compare the performance between methods based on testing 
and methods based on observational study. If the difference is noteworthy, 
it is an interesting topic to investigate the possibility of reducing the gap.

\clearpage
\bibliographystyle{named}
\bibliography{reference}

\end{document}